\pdfoutput=1
\documentclass[11pt]{article}

\usepackage{EMNLP2023}

\usepackage{times}
\usepackage{latexsym}

\usepackage[T1]{fontenc}
\usepackage[utf8]{inputenc}

\usepackage{microtype}

\usepackage{inconsolata}

\usepackage{bbm}
\usepackage{graphicx}
\usepackage{amsmath}
\usepackage{amssymb}
\usepackage{booktabs}
\usepackage{listings}
\usepackage{subcaption}
\usepackage{algorithm}
\usepackage{textcomp}
\usepackage{xcolor}
\usepackage{xspace}
\usepackage{pdfrender}
\usepackage{algpseudocode}
\usepackage{enumitem}
\usepackage{multirow} 
\usepackage[most]{tcolorbox}
\usepackage{pifont}

\title{How to Prompt LLMs for Text-to-SQL: A Study in Zero-shot, Single-domain, and Cross-domain Settings}

\author{Shuaichen Chang \and Eric Fosler-Lussier \\
        The Ohio State University \\ \{chang.1692, fosler-lussier.1\}@osu.edu}

\begin{document}
\maketitle
\newcommand{\eat}[1]{}

\definecolor{pythonblue}{rgb}{0.16,0.12,0.93}
\definecolor{cppgreen}{rgb}{0.16,0.42,0.16}
\definecolor{promptinsert}{HTML}{bfefff}
\definecolor{codehlcolor}{HTML}{ffec8b}
\definecolor{codehlcolor2}{HTML}{ffbbff}
\definecolor{bgcolor}{rgb}{0.95,0.95,0.92}

% Small styles for examples in main text
\lstdefinestyle{plain}{
    basicstyle=\fontsize{8}{7}\ttfamily,
    keywordstyle=\color{blue},
    commentstyle=\color{gray},
    stringstyle=\color{green},
    showstringspaces=false,
    breaklines=true,
    breakatwhitespace=false,
    breakindent=0pt,
    escapeinside={(*@}{@*)}
}

\lstdefinestyle{sql}{
    language=SQL,
    basicstyle=\fontsize{8}{7}\ttfamily,
    keywordstyle=\color{black},
    commentstyle=\color{black},
    stringstyle=\color{black},
    showstringspaces=false,
    breakatwhitespace=false,
    breaklines=true,
    breakindent=0pt,
    escapeinside={(*@}{@*)}
}

\newcommand{\inserthl}[1]{\sethlcolor{promptinsert}\hl{#1}}
\newcommand{\codehl}[1]{\sethlcolor{codehlcolor}\hl{#1}}
\newcommand{\codehlerr}[1]{\sethlcolor{codehlcolor2}\hl{#1}}
\begin{abstract}
Large language models (LLMs) with in-context learning have demonstrated remarkable capability in the text-to-SQL task. Previous research has prompted LLMs with various demonstration-retrieval strategies and intermediate reasoning steps to enhance the performance of LLMs. However, those works often employ varied strategies when constructing the prompt text for text-to-SQL inputs, such as databases and demonstration examples. This leads to a lack of comparability in both the prompt constructions and their primary contributions. Furthermore, selecting an effective prompt construction has emerged as a persistent problem for future research. To address this limitation, we comprehensively investigate the impact of prompt constructions across various settings and provide insights into prompt constructions for future text-to-SQL studies. \footnote{The code for the paper is available at \url{https://github.com/shuaichenchang/prompt-text-to-sql}.}

\end{abstract}

\section{Introduction}

Text-to-SQL models enable users to query databases using natural language questions (NLQs) without having to develop the underlying SQL query.
Over the past few decades, neural models with supervised learning have achieved impressive performance on the text-to-SQL task, which are usually trained on a large training set and then evaluated on test examples \cite{wang2019rat, yu2021grappa, rubin2021smbop, scholak2021picard, gan2021natural,li2023decoupling}. 

Recently, large language models (LLMs) have demonstrated strong capabilities  for in-context learning on many language understanding and generation tasks \cite{brown2020language, chen2021evaluating, chowdhery2022palm}, including on the text-to-SQL task \cite{rajkumar2022evaluating, chang2023dr, liu2023comprehensive}. Instead of training a text-to-SQL model on a large training set, in-context learning allows LLMs to convert a test NLQ into a SQL query using a prompt text. This prompt text includes essential components such as the test database and question. These are accompanied by zero or a few demonstrations:  NLQ-SQL pairs corresponding to either the test database (single-domain) or different databases (cross-domain). Figure \ref{fig:prompt_example} provides an example of a prompt text for a one-shot single-domain task.

\begin{figure}[!tb]
\small
\begin{tcolorbox}[title=Database,left=0mm,right=-2mm,top=-2mm, bottom=-2mm,colback=white]
\begin{lstlisting}[style=sql]
CREATE TABLE Highschooler (
ID int primary key,
name text,
grade int
);
/*
3 example rows:
SELECT * FROM Highschooler LIMIT 3;
ID    name    grade
1510    Jordan    9
1689    Gabriel    9
1381    Tiffany    9
*/
\end{lstlisting}
\end{tcolorbox}
\vspace{-1em}
\begin{tcolorbox}[title=\texttt{Task Instruction},left=0mm,right=-2mm,top=-2mm, bottom=-2mm,colback=white]
\begin{lstlisting}[style=plain]
-- Using valid SQLite, answer the following questions for the tables provided above.
\end{lstlisting}
\end{tcolorbox}
\vspace{-1em}
\begin{tcolorbox}[title=\texttt{Demonstration},left=0mm,right=-2mm,top=-2mm, bottom=-2mm,colback=white]
\begin{lstlisting}[style=sql]
Question: What is Kyle's id?
SELECT ID FROM Highschooler WHERE name = "Kyle";
\end{lstlisting}
\end{tcolorbox}
\vspace{-1em}
\begin{tcolorbox}[title=\texttt{Test Question},left=0mm,right=-2mm,top=-2mm, bottom=-2mm,colback=white]
\begin{lstlisting}[style=plain]
Question: How many high schoolers are there?
SELECT
\end{lstlisting}
\end{tcolorbox}
\vspace{-1em}
\caption{An example of prompt text for 1-shot single-domain text-to-SQL using a snippet of the database \texttt{Network\_1} with a question from the Spider dataset \cite{yu2018spider}.}
\vspace{-2em}
\label{fig:prompt_example}
\end{figure}

Previous research has augmented the text-to-SQL capability of LLMs with demonstration-retrieval strategies \cite{poesia2022synchromesh, shi2022xricl}, intermediate reasoning steps \cite{cheng2022binding,chen2023teaching,pourreza2023din}, and self-debugging ability \cite{chen2023teaching, pourreza2023din}.
However, those studies often employ different prompt strategies that include various
key components of text-to-SQL: database schema and content, and demonstration examples. The difference in prompt constructions makes it difficult to directly compare two studies on their main contribution, and the outcomes of different studies may change based on future revelations in prompt engineering.

In this paper, we evaluate various strategies for prompt construction in three commonly employed text-to-SQL settings: zero-shot, single-domain, and cross-domain. We assess LLMs on text-to-SQL, considering various database prompt constructions in all three settings. Additionally, in the cross-domain scenario, we investigate the strategy for constructing demonstrations. Through our evaluation, we aim to gain insights into the effectiveness of these prompt construction strategies. Our findings can be summarized as follows:
\vspace{-0.4em}
\begin{itemize}[leftmargin=*]
    \setlength\itemsep{-0.4em}
    \item Table relationship and table content play a crucial role in effectively prompting LLMs. However, it is essential to carefully consider their representation in the prompt, as LLMs are sensitive to  the specific presentation in the zero-shot and cross-domain settings.
    \item In-domain demonstration examples can mitigate LLMs' sensitivity to different representations of database knowledge but they cannot replace table content knowledge.
    \item The length of the prompt has a significant impact on the LLMs' performance in the cross-domain setting. We discovered a preferred prompt length that leads to improved performance.
\end{itemize}

\section{In-context Learning for Text-to-SQL}
In the text-to-SQL task, a database and a natural language question (NLQ) are provided as input for generating an output SQL query. Traditional supervised learning approaches train models on specific text-to-SQL datasets. However, in-context learning allows pretrained large language models (LLMs) to 
perform text-to-SQL by providing either zero or a few training examples (NLQ-SQL pairs) as demonstrations. This section introduces three widely used settings for in-context learning in text-to-SQL. Prompt examples in these settings can be found in Appendix \ref{sec:prompt_examples}.

\paragraph{Zero-shot Text-to-SQL} This setting evaluates the text-to-SQL capability of pretrained LLMs to directly infer the NLQ-SQL relationship from a table without any demonstration examples. The input includes a task instruction and a test question with its corresponding database. Zero-shot text-to-SQL is used to directly assess the text-to-SQL capability of LLMs \cite{rajkumar2022evaluating, chang2023dr, liu2023comprehensive}.

\paragraph{Single-domain Few-shot Text-to-SQL} This setting is designed for
applications or domains where it is easy to construct examples, such as booking flights \cite{price1990evaluation,dahl1994atis} and querying geographic information \cite{zelle1996geo}. It tests the ability of LLMs to adapt with a few in-domain demonstration examples, which are collected from the same database as the test question.
The goal is to evaluate how well the LLMs can perform text-to-SQL with minimal in-domain training data \cite{rajkumar2022evaluating}.

\paragraph{Cross-domain Few-shot Text-to-SQL} This setting evaluates the generalization capability of models to new domains by learning from out-of-domain demonstrations. In this scenario, the demonstration NLQ-SQL pairs correspond to one or multiple demonstration databases that are different from the test database. Cross-domain few-shot text-to-SQL assesses how well LLMs can apply their learned knowledge from demonstrations to new databases \cite{poesia2022synchromesh, chen2023teaching}.

\section{Prompt Construction}
A text-to-SQL prompt typically comprises four components: a task instruction, a test database, a test NLQ, and optional demonstrations, as illustrated in Figure \ref{fig:prompt_example}. While the task instruction and test NLQ are easily presented in natural language, there are various strategies for representing the databases and incorporating demonstrations. In this section, we explore different prompt constructions for databases and demonstrations.

\begin{figure}[!tb]
\small
\begin{tcolorbox}[title=\texttt{Table(Columns)}  {\hypersetup{citecolor=white}\cite{liu2023comprehensive}},left=0mm,right=-1mm,top=-2mm, bottom=-2mm,colback=white]
\begin{lstlisting}[style=sql]
Highschooler(ID, name, grade);
Friend(student_id, friend_id);
\end{lstlisting}
\end{tcolorbox}
\vspace{-1em}
\begin{tcolorbox}[title=\texttt{Columns=[]}  {\hypersetup{citecolor=white}\cite{pourreza2023din}},left=0mm,right=-1mm,top=-2mm, bottom=-2mm,colback=white]
\begin{lstlisting}[style=sql]
Table Highschooler, Columns = [ID, name, grade];
Table Friend, Columns = [student_id, friend_id];
\end{lstlisting}
\end{tcolorbox}
\vspace{-1em}
\begin{tcolorbox}[title=\texttt{+FK} {\hypersetup{citecolor=white}\cite{pourreza2023din}},left=0mm,right=-1mm,top=-2mm, bottom=-2mm,colback=white]
\begin{lstlisting}[style=sql]
Foreign_keys = [Friend.student_id = Highschooler.ID, Friend.friend_id = Highschooler.ID];
\end{lstlisting}
\end{tcolorbox}
\vspace{-1em}
\begin{tcolorbox}[title=CreateTable {\hypersetup{citecolor=white}\cite{rajkumar2022evaluating}},left=0mm,right=-1mm,top=-2mm, bottom=-2mm,colback=white]
\begin{lstlisting}[style=sql]
CREATE TABLE Highschooler (
ID int primary key,
name text,
grade int
);
CREATE TABLE Friend (
student_id int,
friend_id int,
primary key (student_id,friend_id),
foreign key(student_id) references Highschooler(ID),
foreign key (friend_id) references Highschooler(ID)
);
\end{lstlisting}
\end{tcolorbox}
\vspace{-1em}
\caption{Examples of the different database schema constructions for a snippet of database \texttt{Network\_1} in Spider.}
\vspace{-1.5em}
\label{fig:prompt_schema}
\end{figure}

\subsection{Database Prompt}
A relational database consists of the database schema and database content. The database schema encompasses the schemas (headers) of tables and the relationship among tables, and database content refers to the data stored in the tables. 

\paragraph{Database Schema}
Figure \ref{fig:prompt_schema} illustrates various prompt constructions for the database schema that have been utilized in previous studies: (1) \texttt{Table(Columns)} \cite{liu2023comprehensive} lists each table along with its columns inside parentheses to represent the table schemas; (2) \texttt{Columns=[]} \cite{pourreza2023din} represents each table along with a list of its columns using an equation-like notation; (3) \texttt{+ForeignKey} \cite{pourreza2023din} further adds foreign keys to indicate the relationships between tables; (4) \texttt{CreateTable} \cite{rajkumar2022evaluating} employed the ``\texttt{Create Table}'' statement to display the table schemas and relationships.

To ensure consistency in the prompt text and accommodate the case-insensitivity of SQL keywords and the database schema, we unify the space and line break in the prompt text and convert all words to lowercase, except for the database content. This normalization process helps to standardize the prompt text. An example is shown in Figure \ref{fig:prompt_norm}.

\paragraph{Database content}
Previous research shows that being aware of database content can improve model performance  by exposing models to the specific format of values in each column \cite{wang2019rat, lin2020bridging,scholak2021picard, rajkumar2022evaluating}. For instance, the phrase ``American student'' could be converted to ``\texttt{WHERE country = `USA'}'' or ``\texttt{WHERE country = `The United States of America'}'' depending on the contents of the \texttt{country} column.

Figure \ref{fig:prompt_content} summarizes different approaches used to construct prompts for showcasing the content of a database.
(1) \texttt{InsertRow} \cite{chen2023teaching}: This method displays $R$ rows of each table by utilizing $R$ ``\texttt{INSERT INTO}'' statements.
(2) \texttt{SelectRow} \cite{rajkumar2022evaluating}: This approach employs the ``\texttt{SELECT * FROM Table LIMIT R}'' query to display the first $R$ rows of each table.
(3) \texttt{SelectCol}:  Instead of presenting table content in a row-wise manner, an alternative method is to use a column-wise format. As there may be duplicated content across different rows, presenting the content column-wise ensures the provision of distinct values within each column to expose LLMs to a broader range of content. We propose using the query ``\texttt{SELECT DISTINCT [Column] FROM [Table] LIMIT R}'' to list $R$ distinct cell values in each column.

\begin{figure}[!tb]
\small
\begin{tcolorbox}[title=\texttt{InsertRow} \hypersetup{citecolor=white}\cite{chen2023teaching},left=0mm,right=-1mm,top=-2mm, bottom=-2mm,colback=white]
\begin{lstlisting}[style=sql]
INSERT INTO Highschooler (ID, name, grade) VALUES (1510, "Jordan", 9);
INSERT INTO Highschooler (ID, name, grade) VALUES (1689, "Gabriel", 9);
INSERT INTO Highschooler (ID, name, grade) VALUES (1381, "Tiffany", 9);\end{lstlisting}
\end{tcolorbox}
\vspace{-1em}
\begin{tcolorbox}[title=\texttt{SelectRow} \hypersetup{citecolor=white}\cite{rajkumar2022evaluating},left=0mm,right=-1mm,top=-2mm, bottom=-2mm,colback=white]
\begin{lstlisting}[style=sql]
/*
3 example rows:
SELECT * FROM Highschooler LIMIT 3;
ID    name    grade
1510    Jordan    9
1689    Gabriel    9
1381    Tiffany    9
*/
\end{lstlisting}
\end{tcolorbox}
\vspace{-1em}
\begin{tcolorbox}[title=SelectCol (Ours), left=0mm,right=-1mm,top=-2mm, bottom=-2mm,colback=white]
\begin{lstlisting}[style=sql]
/*
Columns in Highschooler and 3 distinct examples in each column:
ID: 1025, 1101, 1247
name: "Jordan", "Gabriel", "Tiffany"
grade: 9, 10, 11
*/
\end{lstlisting}
\end{tcolorbox}
\vspace{-1em}
\caption{Examples of the different database content constructions for showing 3 cell values in each column for the \texttt{Highschool} table in Figure \ref{fig:prompt_schema}.}
\label{fig:prompt_content}
\vspace{-2em}
\end{figure}

\subsection{Demonstration Prompt}
In few-shot settings, LLMs are provided with demonstrations within the prompt text. In the single-domain few-shot setting, we incorporate a few pairs of NLQs and SQLs as demonstrations inserted between the test database and question, following previous work \cite{rajkumar2022evaluating}. In the cross-domain few-shot setting, we use both out-of-domain NLQ-SQL pairs (demonstration examples) and corresponding databases (demonstration databases)  placed before the test database and question. Prior research in the $N$-shot setting either uses one demonstration database with $N$ examples \cite{pourreza2023din} or employs $N$ demonstration databases, each with a single NLQ-SQL pair \cite{poesia2022synchromesh, chen2023teaching}. In contrast, we consider a more general scenario where the demonstrations comprise $M$ databases, each with $K$ NLQ-SQL pairs, with $M \times K=N$. We list the examples of 4-shot single-domain and cross-domain demonstrations in Appendix \ref{sec:prompt_examples}.

Additionally, we normalize demonstration SQL queries by first parsing the SQL queries and unifying their format, such as using lowercase for SQL keywords and database schema and unifying the space around punctuation. Figure \ref{fig:prompt_norm} provides an example of SQL normalization.

\begin{figure}[!tb]
\small
\begin{tcolorbox}[title=\texttt{Unnormalized database and SQL},left=0mm,right=-2mm,top=-2mm, bottom=-2mm,colback=white]
\begin{lstlisting}[style=sql]
-- Database Schema
CREATE TABLE Highschooler(
	ID int primary key, 
	name text, 
	grade int);
 
-- SQL Query
SELECT count( * ) FROM Highschooler WHERE Name = "Kyle";
\end{lstlisting}
\end{tcolorbox}
\vspace{-1em}
\begin{tcolorbox}[title=\texttt{Normalized database and SQL},left=0mm,right=-2mm,top=-2mm, bottom=-2mm,colback=white]
\begin{lstlisting}[style=sql]
-- Database Schema
create table highschooler (
id int primary key,
name text,
grade int
);

-- SQL Query
select count(*) from highschooler where name = 'Kyle';
\end{lstlisting}
\end{tcolorbox}
\vspace{-1em}
\caption{An example of the normalization for database and SQL prompts.}
\label{fig:prompt_norm}
\vspace{-2em}
\end{figure}

\section{Experiments}

\paragraph{Data \& Evaluation} For our experiments, we utilize the Spider dataset \cite{yu2018spider}, a cross-domain benchmark for the text-to-SQL task. We conduct our experiments on the development set of Spider (Spider-dev) as the test set is not publicly available. Spider-dev consists of 20 databases with 1034 pairs of NLQ and SQL in total. We evaluate models with execution accuracy (EX) which compares the execution results of a predicted SQL and a gold SQL.

In the cross-domain setting, we use the training set of Spider to select demonstration examples. As a few databases contain long schema that may cause the prompt to exceed the token limits of LLMs, we only use the databases with fewer than 1000 tokens when constructing the \texttt{CreateTable} prompt. This results in a total of 130 databases being used as demonstration databases in the cross-domain setting.

\paragraph{Models}  We used GPT-3 Codex \cite{chen2021evaluating} and ChatGPT due to their demonstrated performance and prevalence in the field.\footnote{We employ the Code-davinci-002 version of Codex across all settings. In zero-shot and single-domain setups, we utilize the \texttt{gpt-3.5-turbo-0301} version of ChatGPT. For cross-domain experiments involving ChatGPT-16K, we turned to \texttt{gpt-3.5-turbo-16k-0613} due to its extended context length.}

\paragraph{Experiment Setup}
For the zero-shot setting, we construct each prompt text with a task instruction, a test database, and a test question. We include $R=3$ table rows in the database prompt,
which has been discovered as the optimal number in previous work \cite{rajkumar2022evaluating}. For the few-shot settings, we incorporate $N$ demonstration examples in addition to the zero-shot prompt text. 

In the single-domain text-to-SQL scenario, we use a leave-one-out split, as some databases in Spider-dev contain a small number of examples. When evaluating one example, we regard all other examples from the same database as the training set and randomly retrieve $N$ examples from them. Since Spider contains multiple NLQs corresponding to the same SQL query, we require that the training set does not contain examples with the same SQL template as the test example, again following  previous work \cite{finegan2018improving}.

In the cross-domain scenario, we randomly select $M$ demonstration databases, each with $K$ NLQ-SQL pairs ($M\times K =N$) from the Spider training set. 
Incorporating multiple demonstration databases in a prompt text significantly increases its length. Hence, we only use Codex for the cross-domain experiments, due to its higher token limit of 8K, surpassing the 4K limit of ChatGPT.
In both single-domain and cross-domain settings, we compare different prompt construction methods using the same few-shot examples to make a fair comparison. We repeat our experiments three times and present the average results.

\begin{table*}[!htb]
    \centering
    \small
    \begin{tabular}{llcccc}
    \toprule
    & & \multicolumn{2}{c}{\textbf{Codex}} & \multicolumn{2}{c}{\textbf{ChatGPT}} \\
    \midrule
     \multicolumn{2}{c}{\textbf{Database Prompt Construction}} & \bf \# Tokens (U|N) & \bf EX (U|N) & \bf \# Tokens (U|N) & \bf EX (U|N)\\
    \midrule
    \multirow{2}{*}{\texttt{Table Schema}} & \texttt{Table(Columns)}   & 148 | \underline{147} & 69.0 | \underline{71.9} & 118 | \underline{115} & 68.8 | \underline{70.5} \\
    \cmidrule{2-6}
    & \texttt{Columns=[]}  &  169 | \underline{167} &  70.2 | \underline{71.8} &  137 | \underline{135} &  68.3 | \underline{69.1}\\
    \midrule
    \midrule
    \multirow{2}{*}{\texttt{+Relationship}} & \texttt{Columns=[]+ForeignKey} & 226 | \underline{223} & 72.3 | \underline{73.1} & 178 | \underline{174} & \underline{72.9} | 71.2 \\
    \cmidrule{2-6}
    & \texttt{CreateTable}  & 474 | \underline{356} & 71.8 | \underline{73.1} & 339 | \underline{254} & 70.7 | \underline{71.7}\\
    \midrule
    \midrule
    \multirow{2}{*}{\texttt{+Relationship+Content}} & \texttt{CreateTable+InsertRow 3} & 1089 | \underline{1013} & 70.9 | \underline{71.9} & 964 | \underline{872} & \underline{71.8} | \underline{71.8}\\
    \cmidrule{2-6}
    & \texttt{CreateTable+SelectRow 3}  & 820 | \underline{770} &  73.3 | \underline{74.1} & 761 | \underline{674} &  71.8 | \underline{72.1} \\
    \cmidrule{2-6}
    & \texttt{CreateTable+SelectCol 3}  & 958 | \underline{831} & \textbf{75.0} | \textbf{\underline{75.7}}  & 799 | \underline{712} &  \textbf{73.3} | \textbf{\underline{73.6}} \\
    
    \bottomrule
    \end{tabular}
    \caption{Zero-shot results of Codex and ChatGPT using different database prompt constructions. \texttt{Table Schema} (upper part) contains prompts that solely include the schema of tables, while \texttt{+Relationship} (middle part) incorporates foreign keys as the table relationships and \texttt{+Relationship+Content} (lower part) adds table content as well. \# Tokens is the average token counts in the prompts and EX represents the execution accuracy of SQLs. U|N represents the results of unnormalized prompts and normalized prompts, respectively. The underlines highlight the lower number of tokens and higher accuracies when comparing unnormalized and normalized prompts and the highest accuracy achieved among all prompts is highlighted in bold.
    }
    \vspace{-1em}
    \label{tab:schema_prompt_results}
\end{table*}

\section{Results}

In this section, we present our empirical findings in the areas of zero-shot, single-domain, and cross-domain text-to-SQL.
Through our experiments, we aim to answer a few crucial research questions in each setting and provide insightful strategies for future studies on effective prompting.

\subsection{Zero-shot Text-to-SQL}

In the zero-shot setting, we focus on comparing different prompt constructions for databases. Table \ref{tab:schema_prompt_results} shows the average prompt length and execution accuracy of Codex and ChatGPT using various database prompt constructions.

\noindent \textbf{Q1: How does normalized database prompt perform compared to unnormalized ones?} Normalized schemas are found to have a reduced token count in comparison to unnormalized schemas across all database constructions. The normalization also tends to yield slightly better performance. As for Codex, normalized schemas show improvement in all prompts. For ChatGPT, normalized schemas either improve accuracy or achieve the same accuracy or achieve the same level of accuracy as unnormalized schemas in 6 out of 7 schema constructions. The tests of statistical significance are presented in Appendix \ref{sec:significance}.

\noindent \textbf{Q2: What database knowledge is crucial for effectively prompting LLMs?} 
Our experiments indicate that table relationships and content are important. The \texttt{Columns=[]} prompt includes only the table schema, while the \texttt{Columns=[]+ForeignKey} prompt contains the additional relationship among tables shown as foreign keys. Including such information improves the performance of both Codex (71.8 -> 73.1) and ChatGPT (69.1 -> 71.2).
Moveover, exposing LLMs to database content with the \texttt{SelectRow} and \texttt{SelectCol} prompts further enhances the performance of both Codex and ChatGPT, while the \texttt{InsertRow} prompt does not seem to be beneficial. We believe that database content is valuable, but its representation needs to be carefully chosen.

\noindent \textbf{Q3: How does Codex perform compared to ChatGPT?} 
While we do not focus on comparing different LLMs on the text-to-SQL tasks in this paper, it is worth noting that Codex consistently outperforms ChatGPT on zero-shot text-to-SQL using various prompt constructions. 

Based on all the findings above, we would recommend using Codex in conjunction with normalized \texttt{CreateTableSelectCol} prompt construction for zero-shot text-to-SQL.\footnote{To simplify our experiments and ensure consistent prompts, we adopt normalization for single-domain and cross-domain experiments.}

\subsection{Single-domain Text-to-SQL}
In the zero-shot text-to-SQL setting, we discovered that the prompt constructions of databases impact the performance of LLMs. This discovery naturally raises the question of whether the introduction of in-domain demonstrations affects the performance of LLMs to different database prompts.

\begin{figure}[!htb]
    \centering
    \subcaptionbox{Codex}
    {\includegraphics[width=0.95\columnwidth]{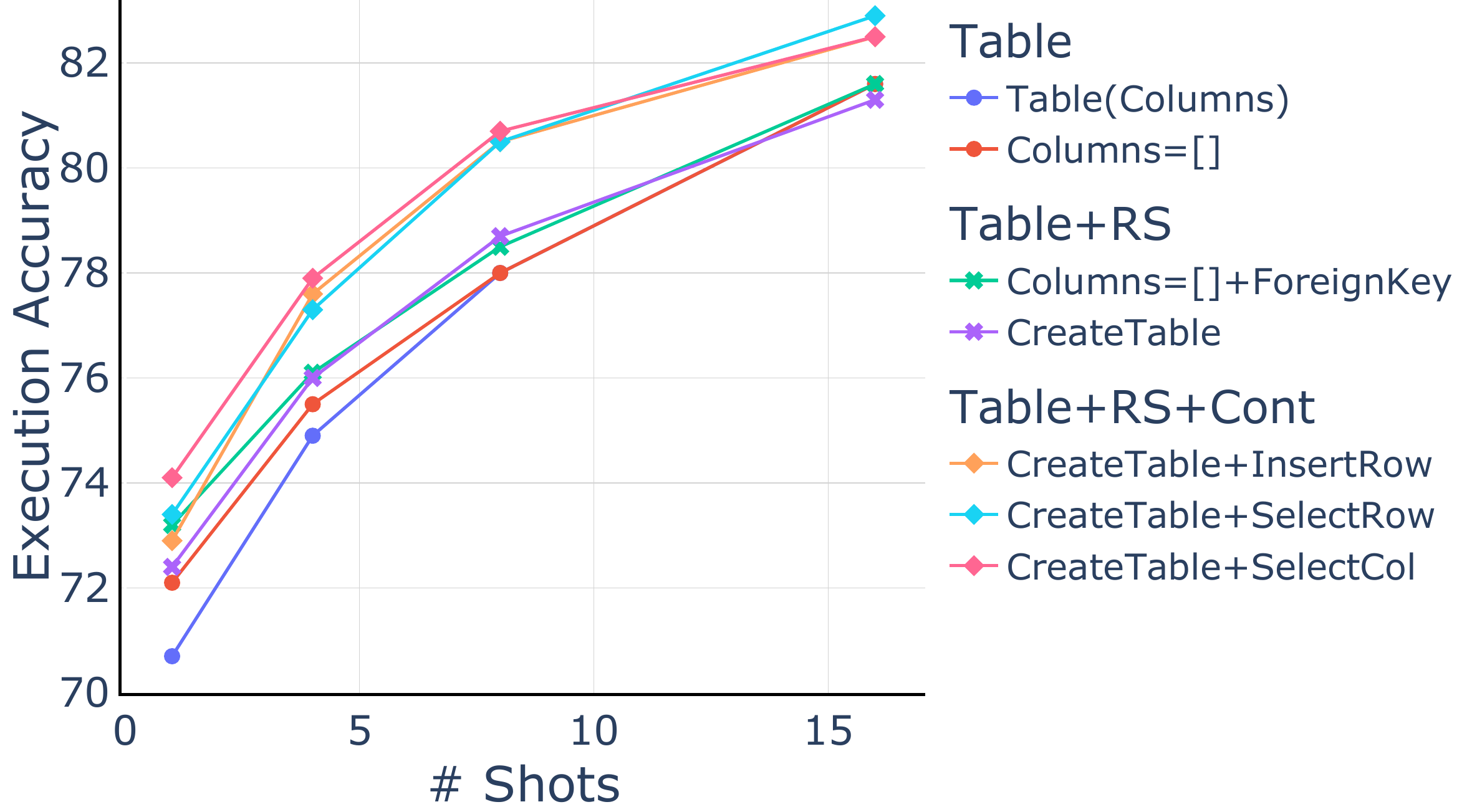}  }
    \subcaptionbox{ChatGPT}
    {\includegraphics[width=0.95\columnwidth]{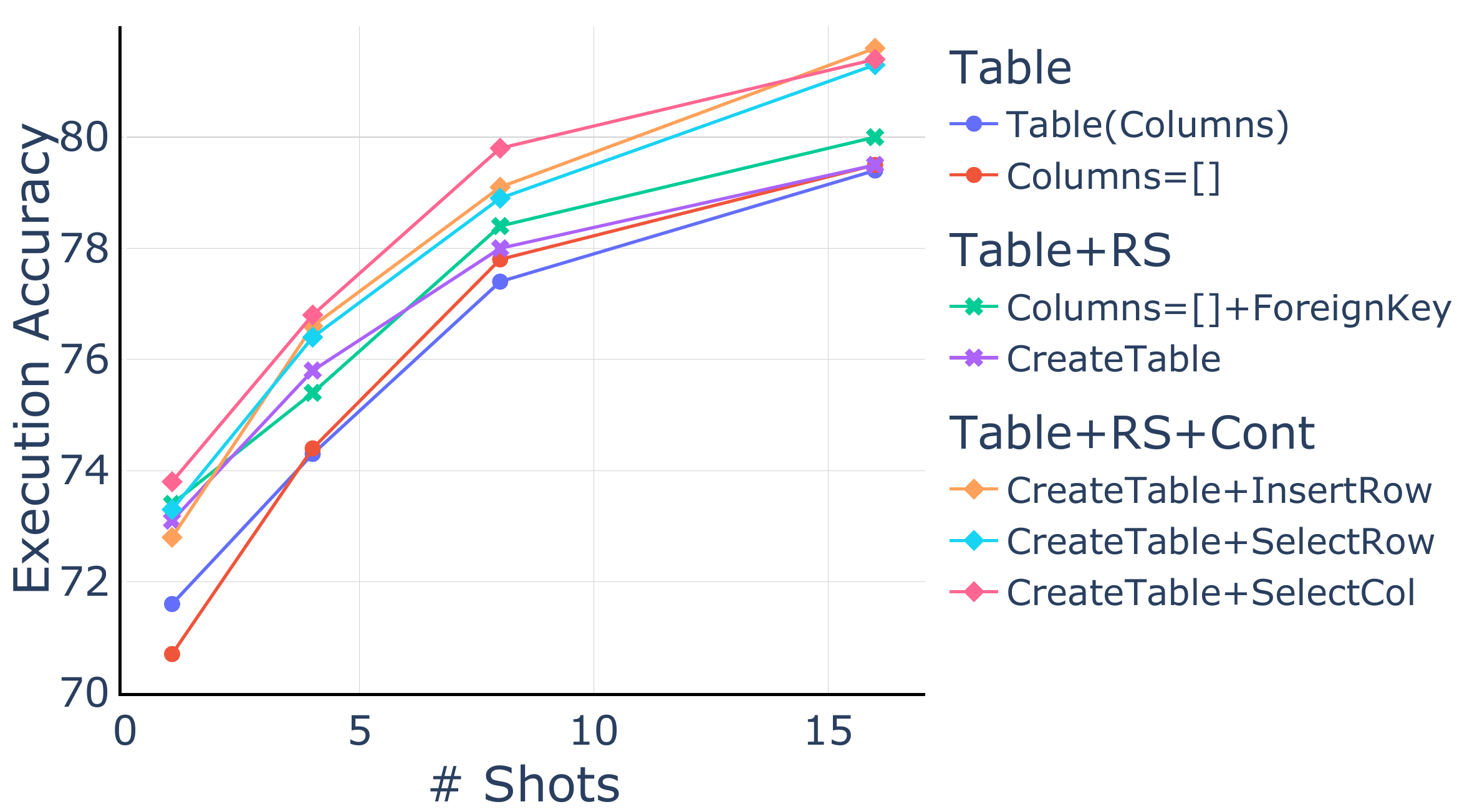}  }  
    \caption{Execution accuracy of Codex and ChatGPT for single-domain text-to-SQL with 1, 4, 8, and 16 in-domain examples. \texttt{RS} and \texttt{Cont} correspond to table relationship and table content, respectively. Detailed results can be found in Table \ref{tab:single_domain_codex_results} and \ref{tab:single_domain_chatgpt_results}.}
    \label{fig:single_domain_results}
    \vspace{-1em}
\end{figure}

\noindent \textbf{Q1: Does the use of in-domain demonstrations enhance LLM's performance?} Figure \ref{fig:single_domain_results} depicts the performance of Codex and ChatGPT using different database prompt constructions with respect to different numbers of in-domain demonstration examples. For all database prompts, the performance of LLMs experiences a notable improvement when in-domain examples are presented. Furthermore, the performance continues to enhance as the number of in-domain examples increases.

\noindent \textbf{Q2: What database knowledge is important when presenting in-domain demonstrations?} 
While we have observed that the presence of table relationships and table content enhanced LLMs' performance in the zero-shot scenario, it is not clear whether they are still important in the single-domain setting. A hypothesis is that table relationship and table content knowledge can be acquired from in-domain examples as they may appear in SQL clauses \texttt{JOIN} and \texttt{WHERE}.

For table relationships, we compare two database prompt constructions \texttt{Columns=[]} and \texttt{Columns=[]+ForeignKey}. Both construct the table schema in the same way while the latter includes foreign keys as table relationships. In the zero-shot scenario, \texttt{Columns=[]+ForeignKey} outperforms \texttt{Columns=[]} by 1.3 and 2.1 for Codex and ChatGPT, respectively. However, as increasing the number of in-domain examples, we notice a gradual reduction in the performance gap between these two prompts. With the utilization of 16 in-domain examples, the gap completely disappears for Codex, while ChatGPT exhibits a marginal difference of only 0.5\%. 

For table content, we compare \texttt{CreateTable} with \texttt{CreateTable+SelectCol}. Both contain the same prompts for presenting the table schema and relationship,  while the latter additionally includes table content.
In the zero-shot scenario, \texttt{CreateTable+SelectCol} outperforms \texttt{CreateTable} by 2.0\% for Codex and 1.7\% for ChatGPT. As we proceed to increase the number of in-domain examples, we observe that the performance gap between these two prompts does not exhibit a significant reduction. Even with 16 in-domain examples, the gap still persists at 1.3 for Codex and 1.9 for ChatGPT.

These results indicate LLMs are able to quickly learn table relationships from a small number of in-domain demonstrations, however, it is more challenging to obtain table content knowledge from demonstration examples. Consequently, the inclusion of table content remains crucial for achieving satisfactory performance in the single-domain text-to-SQL scenario.

\noindent \textbf{Q3: Can in-domain demonstrations alleviate the sensitivity of LLMs to the representation of table content?}
In the zero-shot setting, we observe that LLMs are sensitive to how the table content is presented. Specifically, \texttt{SelectCol 3} outperforms \texttt{InsertRow 3} by a substantial margin of 3.8 for Codex and 1.8 for ChatGPT.
However, as we expose LLMs to in-domain demonstrations, LLMs become less sensitive to the specific representation of table content. The performance disparities among the three table content prompts become marginal. Notably, with only 4 examples, the performance difference between \texttt{SelectCol 3} and \texttt{InsertRow 3} diminishes to 0.3 for Codex and 0.2 for ChatGPT.

To summarize, in single-domain text-to-SQL, we recommend incorporating a greater number of in-domain examples whenever feasible. It is also essential to ensure the presence of table content in conjunction with the table schema while the specific choice of table content construction is less crucial compared to the zero-shot scenario.

\subsection{Cross-domain Text-to-SQL}

In this section, we present the results to answer a series of questions regarding the demonstration and database prompt construction. 

\subsubsection{Impact of Demonstration Prompt}
To investigate the impact of the number of databases and examples per database in demonstrations, we conduct experiments encompassing various combinations. Specifically, our demonstrations are composed of M demonstration databases, each containing K NLQ-SQL pairs.  We consider scenarios with up to 8 databases and 16 examples per database as long as the combination does not exceed the prompt length limit. We opt to use the database prompt \texttt{CreateTable+SelectRow 3} as it contains fewer tokens compared to \texttt{InsertRow} and \texttt{SelectCol} while encompassing all valuable database knowledge. We present the experiments with Codex in this section. Experiments involving ChatGPT-16K can be found in Appendix \ref{sec:appendix:chatgpt-16k} which show similar results as Codex.

\begin{figure}[!tb]
  \centering
  \includegraphics[width=.95\linewidth]{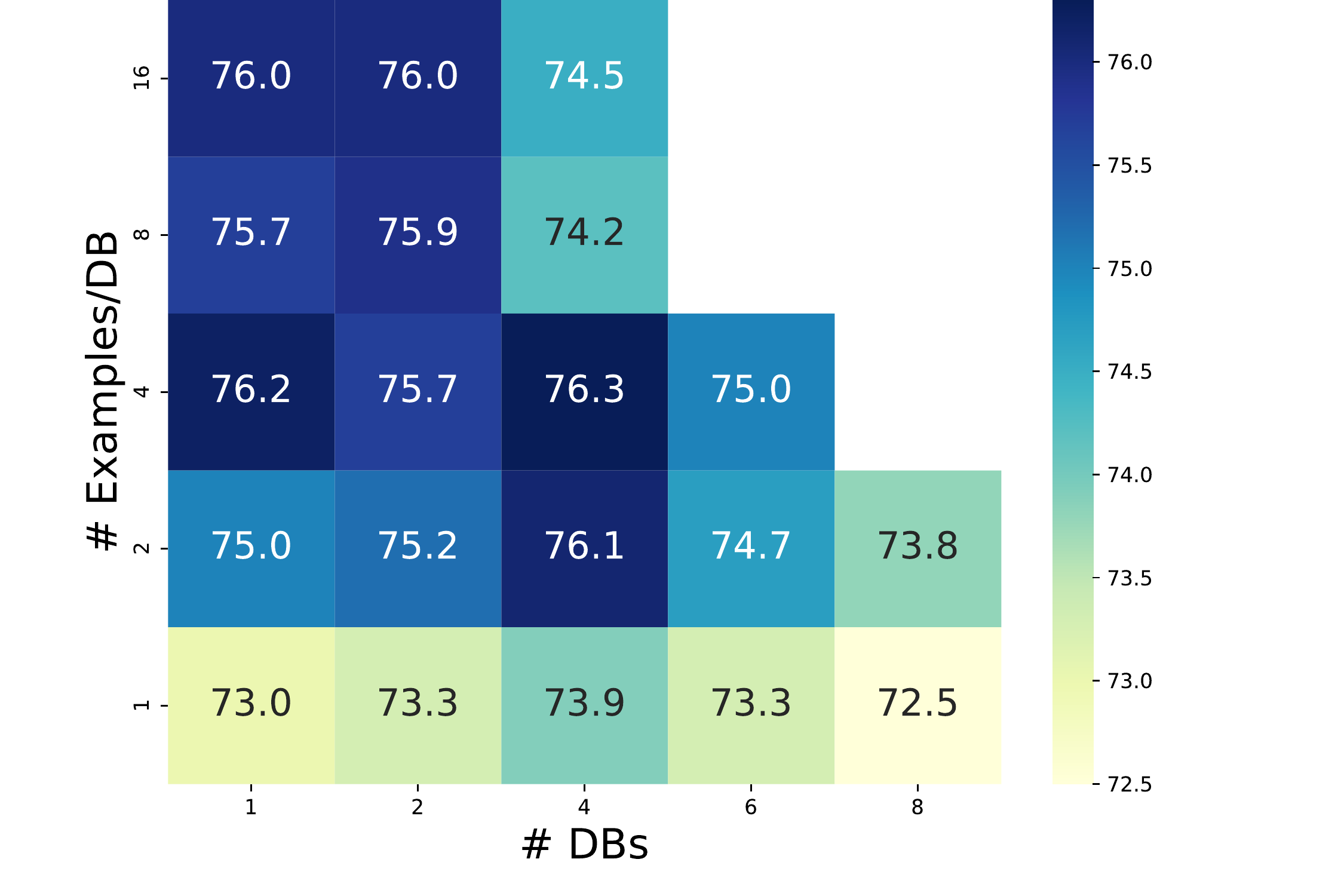}
  \caption{A heat map of Codex's execution accuracy using \texttt{CreateTable+SelectRow 3} for different numbers of databases and examples per database in the demonstration. Darker color indicates higher accuracy.}
  \label{fig:cross_domain_results}
  \vspace{-1em}
\end{figure}
\begin{figure}[!tb]
  \centering
  \includegraphics[width=.95\linewidth]{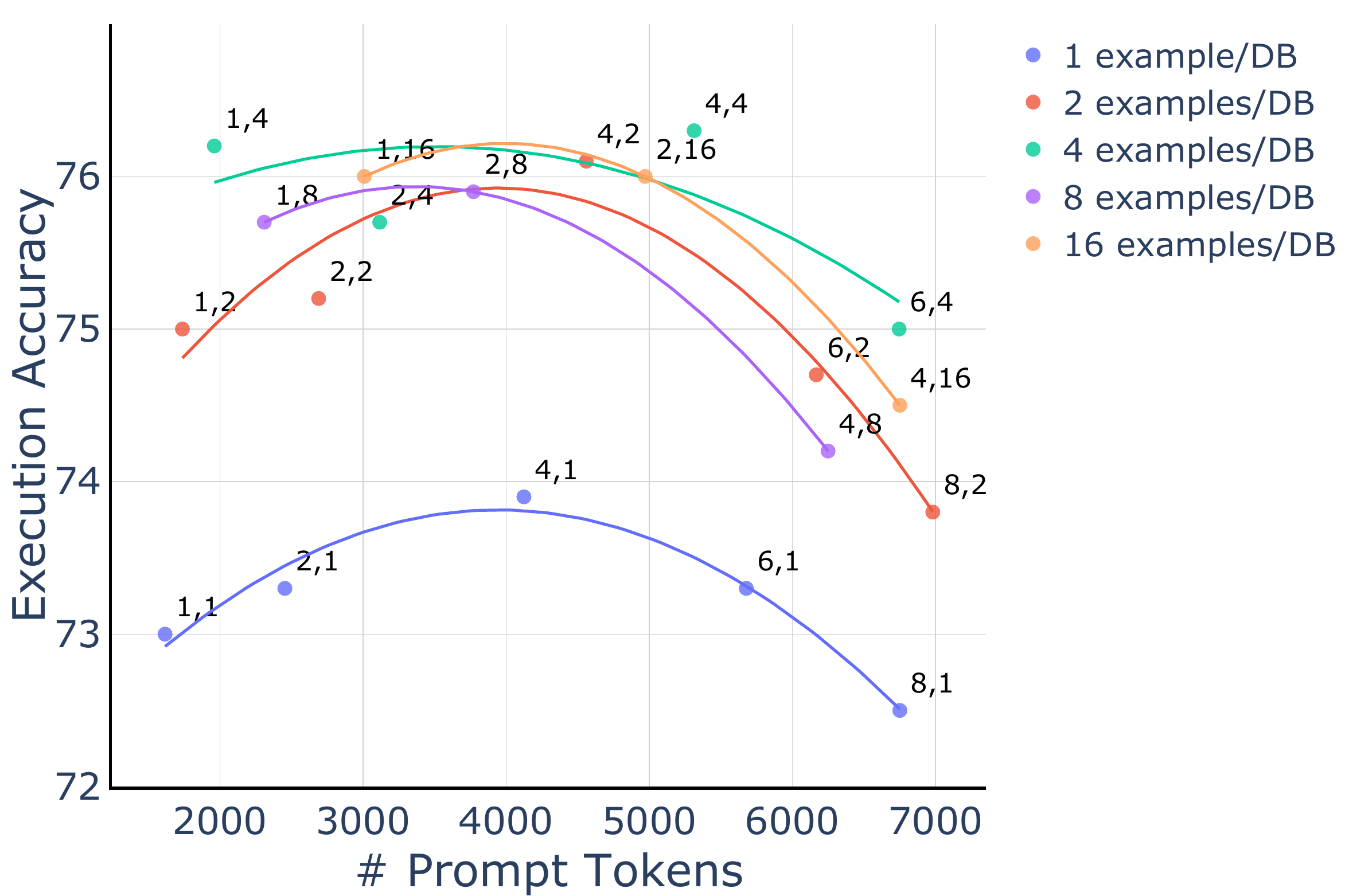}
  \caption{Execution accuracy of Codex in relation to the length of prompts. Each dot on the graph represents a specific demonstration prompt construction, with the $m, k$ denoting the number of databases and examples per database used in the prompt. The lines represent second-degree polynomial trendlines fitted to the results.}
  \label{fig:cross_domain_results_prompt_len}
\vspace{-1.5em}
\end{figure}

\noindent \textbf{Q1: Does increasing demonstration examples enhance LLMs' performance?} 
Figure \ref{tab:cross_domain_codex_results} presents the accuracy of Codex corresponding to different combinations of the number of databases and the number of examples per database used as demonstrations.
We analyze the results from two perspectives. Firstly, for a fixed number of databases, we observe an initial improvement in Codex's performance as the number of examples per database increases. However, this improvement plateaus or declines once 4 examples per database are provided. Surprisingly, when using 4 databases, employing 8 or 16 examples per database leads to a significant decrease in the Codex's performance compared to using 2 or 4 examples per database.
Secondly, for a fixed number of examples per database, we observe an initial increase in Codex's performance as the number of databases increases, however, this improvement is followed by a significant decrease once the number of databases reaches a certain threshold (either 4 or 6). 

\noindent \textbf{Q2: Why does increasing the number of databases decrease LLMs' performance?} 
As depicted in Figure \ref{tab:cross_domain_codex_results}, presenting more databases does not always lead to improved performance. In fact, there is a significant decline in performance, once it surpasses a threshold. We hypothesize that this phenomenon is attributed to the length of the prompt text. To test this hypothesis, we analyze the results in relation to the prompt length.

Figure \ref{fig:cross_domain_results_prompt_len} shows the relationship between the accuracy of different demonstration prompts and their prompt lengths. Notably, the performance of Codex exhibits an inverted-U shape as the prompt length increases for each number of examples per database. Additionally, we observe a substantial drop in performance once the prompt text length exceeds approximately 5500 tokens. Similarly, Figure \ref{fig:cross_domain_results_prompt_len_chatgpt16k} shows that the performance of ChatGPT-16K starts to decrease when prompt text length exceeds 11K tokens.
Based on these observations, we conjecture that LLMs may have a sweet spot in terms of prompt length, potentially influenced by factors such as their model architecture or training data. This indicates that even though LLMs are capable of handling long contexts, they may not necessarily perform better with excessively long prompts.

\begin{table*}
    \centering
    \small
    \begin{tabular}{llcccccc}
    \toprule
     \multicolumn{2}{c}{\textbf{Database Prompt Construction}} & \bf 0-shot & \bf 1-shot & \bf 2-shot & \bf 4-shot & \bf 8-shot & \bf 16-shot \\
    \midrule
    \multirow{2}{*}{\texttt{Table Schema}} & \texttt{Table(Columns)} & 71.9 & 72.0 & 73.0 & 73.2 & 72.8 & 73.9 \\
    \cmidrule{2-8}
    & \texttt{Columns=[]} &71.8 & 71.9 & 73.6 & 74.2 & 73.7 & 74.4 \\
    \midrule
    \midrule
    \multirow{2}{*}{\texttt{+Relationship}} & \texttt{Columns=[]+ForeignKey} & 73.1 & \underline{73.3} & 74.5 & 74.9 & 74.9 & 75.2 \\
    \cmidrule{2-8}
    & \texttt{CreateTable} & 73.1 & 72.1 & 73.4& 73.7 & 74.1 & 75.1 \\
    \midrule
    \midrule
    \multirow{2}{*}{\texttt{+Relationship+Content}} & \texttt{CreateTable+InsertRow 3} &71.9 &72.2 & 74.1 & 74.9 & 74.9 & 74.8\\
    \cmidrule{2-8}
    & \texttt{CreateTable+SelectRow 3} & \underline{74.1} & 73.0 & \underline{75.0} &\underline{76.2} & \underline{75.7} & \underline{76.0}\\
    \cmidrule{2-8}
    & \texttt{CreateTable+SelectCol 3} & \textbf{75.7} & \textbf{74.4} & \textbf{75.5} & \textbf{76.5} & \textbf{76.8} & \textbf{76.5}\\
    \bottomrule
    \end{tabular}
    \caption{Cross-domain results of Codex using different database prompt constructions. Only one demonstration database is included in a prompt, N-shot represents N examples corresponding to the demonstration database. The best and second-best results for each shot are highlighted in bold and underlined.}
    \label{tab:cross_domain_codex_results}
    \vspace{-1em}
\end{table*}

\subsubsection{Impact of Database Prompt}
Since incorporating demonstration databases may cause a decrease in Codex's performance, we focus our database prompt experiments on using one demonstration database in combination with varying quantities of demonstration examples. Table \ref{tab:cross_domain_codex_results} presents the execution accuracy of Codex using different database prompts.

\noindent \textbf{Q3: Do different database prompts show similar trends with the number of demonstration examples?} 
We observe an initial performance increase for all database prompts. However, once more than 4 examples are provided, the improvement starts to level off, indicating that the different database prompts exhibit similar trends in relation to the number of demonstration examples.

\noindent \textbf{Q4: Can out-of-domain demonstrations alleviate the sensitivity of LLMs to database prompts?} 
First, we observe that incorporating table relationships and content in the prompts remains crucial for effectively prompting Codex in the cross-domain setting. This is not surprising, as Codex cannot directly learn knowledge specific to the test database from the out-of-domain demonstrations.
Furthermore, we find that Codex continues to exhibit sensitivity to the representation of table content. Despite having demonstration databases that mirror the construction of the test database, Codex still displays a preference for\texttt{SelectRow} and \texttt{SelectCol} when presenting table content, compared to \texttt{InsertCol}.

In conclusion, while out-of-domain demonstrations enhance LLMs' capabilities in text-to-SQL, they do not provide database-specific knowledge. Consequently, careful construction of database prompts remains crucial, aligning with the observations made in the zero-shot setting.
\section{Related Work}

\paragraph{LLMs for Text-to-SQL}
In recent years, there has been significant progress in leveraging LLMs for the text-to-SQL task. Various methods have been proposed to enhance the capabilities of LLMs. For example, \citet{rubin2021learning, poesia2022synchromesh} have demonstrated the effectiveness of similarity-based demonstration retrieval in the cross-domain setting. Additionally, \citet{levy2022diverse} have highlighted the advantages of incorporating diverse demonstrations for compositional generalization.
Furthermore, \citet{pourreza2023din} and \citet{chen2023teaching} incorporate intermediate steps in prompts and unlock LLMs' capability of self-correcting their predictions. 

In contrast to these approaches, our focus lies in conducting a comprehensive evaluation of prompt representations across different text-to-SQL settings. While there are similar motivations to the work by \citet{rajkumar2022evaluating}, which analyzes the performance of CodeX on Spider for the zero-shot setting and on two databases for the single-domain setting, we aim to provide more general findings by evaluating across a wider range of databases and considering all three text-to-SQL settings.

\paragraph{Table Representation}
Encoding structured databases with neural models has been a persistent challenge. To encode database schema, graph neural networks are utilized to represent the relationships among tables \cite{bogin2019representing, chen2021shadowgnn}. 
Alternatively, other studies \cite{guo2019towards,lin2020bridging,shaw2020compositional} have converted table schemas into a sequence to effectively leverage pretrained language models, such as BERT \cite{devlin2018bert} and T5 \cite{raffel2020exploring}. In such cases, table relationships can be encoded as meta-data features \cite{lin2020bridging} or used as a guide for attention mechanism \cite{wang2019rat, cao2021lgesql, li2023graphix}.

To incorporate table content into neural models, prior supervised methods provide question-specific table content by identifying the relevant table content mentioned in the question through string matching \cite{lin2020bridging, shaw2020compositional}. However, \citet{chang2023dr} have revealed the vulnerability of string matching to perturbations. Given that LLMs with in-context learning support longer input sequences compared to supervised methods, we follow previous work to provide table content without explicitly considering the questions \cite{rajkumar2022evaluating,chen2023teaching}.
% \vspace{-0.5em}
\section{Conclusions}
% \vspace{-0.5em}
In this paper, we investigate effective prompting strategies in the text-to-SQL task. We thoroughly compare various prompt construction strategies for databases and demonstrations in the zero-shot, single-domain, and cross-domain text-to-SQL. Through our investigation, we uncover the critical database knowledge and optimal representations for effective prompting. Additionally, an interesting finding is the existence of a sweet spot in terms of prompt length for Codex in the cross-domain setting. Overall, we believe that our findings will provide valuable guidance for future research in the field of text-to-SQL with LLMs.

\section*{Limitation}
We conducted our experiments using 20 databases from the Spider dataset, with the goal of providing general findings for text-to-SQL prompt constructions. However, our findings may not always be applicable to a specific database, particularly if the database is significantly different from the Spider databases.
For the single-domain and cross-domain text-to-SQL scenarios, we conduct our experiments multiple times, each involving randomly selecting demonstrations with different random seeds, however, we did not investigate the effectiveness of prompt constructions with different demonstration-retrieval strategies or intermediate reasoning steps.

\section*{Ethics Statement}
We acknowledge the importance of the ACL Ethics Policy and agree with it. In this paper, we use OpenAI Codex and ChatGPT as our language models \footnote{API is available at https://openai.com/api/.}. Codex is currently free for research purposes, the cost of ChatGPT is around \$200. 
The code for the paper is included in the supplementary materials and will be publicly released to facilitate reproducibility.

\newpage
\bibliography{anthology,custom}
\bibliographystyle{acl_natbib}

\clearpage
\appendix
\section{Appendix}

\subsection{Prompt Examples} 
\label{sec:prompt_examples}
Below contains an example of a zero-shot normalized prompt, which contains the database \texttt{Network\_1} from Spider \cite{yu2018spider}, a task instruction ``Using valid SQLite, answer the following questions for the tables provided above.'', and a test question ``How many high schoolers are there?''. 

\begin{tcolorbox}[title={Zero-shot normalized prompt},breakable, toprule at break=0pt, bottomrule at break=0pt,left=0mm,right=-1mm,top=-1mm, bottom=-1mm,colback=white]
\begin{lstlisting}[style=sql]
create table highschooler (
id int primary key,
name text,
grade int
);
/*
3 example rows:
select * from highschooler limit 3;
id    name    grade
1510    Jordan    9
1689    Gabriel    9
1381    Tiffany    9
*/

create table friend (
student_id int,
friend_id int,
primary key (student_id,friend_id),
foreign key(student_id) references highschooler(id),
foreign key (friend_id) references highschooler(id)
);
/*
3 example rows:
select * from friend limit 3;
student_id    friend_id
1510    1381
1510    1689
1689    1709
*/

create table likes (
student_id int,
liked_id int,
primary key (student_id, liked_id),
foreign key (liked_id) references highschooler(id),
foreign key (student_id) references highschooler(id)
);
/*
3 example rows:
select * from likes limit 3;
student_id    liked_id
1689    1709
1709    1689
1782    1709
*/

-- Using valid SQLite, answer the following questions for the tables provided above.

Question: How many high schoolers are there?
select
\end{lstlisting}
\end{tcolorbox}

\newpage

Below contains an example of a 4-shot single-domain normalized prompt, which contains a database prompt and 4 demonstration examples ahead of the test question. 

\begin{tcolorbox}[title={4-shot single-domain normalized prompt},breakable, toprule at break=0pt, bottomrule at break=0pt,left=0mm,right=-1mm,top=-1mm, bottom=-1mm,colback=white]
\begin{lstlisting}[style=sql]
create table highschooler (
id int primary key,
name text,
grade int
);
/*
3 example rows:
select * from highschooler limit 3;
id    name    grade
1510    Jordan    9
1689    Gabriel    9
1381    Tiffany    9
*/

create table friend (
student_id int,
friend_id int,
primary key (student_id,friend_id),
foreign key(student_id) references highschooler(id),
foreign key (friend_id) references highschooler(id)
);
/*
3 example rows:
select * from friend limit 3;
student_id    friend_id
1510    1381
1510    1689
1689    1709
*/

create table likes (
student_id int,
liked_id int,
primary key (student_id, liked_id),
foreign key (liked_id) references highschooler(id),
foreign key (student_id) references highschooler(id)
);
/*
3 example rows:
select * from likes limit 3;
student_id    liked_id
1689    1709
1709    1689
1782    1709
*/

-- Using valid SQLite, answer the following questions for the tables provided above.

Question: What is Kyle's id?
select id from highschooler where name = 'Kyle';
Question: Return the names of friends of the high school student Kyle.
select t3.name from friend as t1 join highschooler as t2 on t1.student_id = t2.id join highschooler as t3 on t1.friend_id = t3.id where t2.name = 'Kyle';
Question: Show names of all high school students who do not have any friends.
select name from highschooler except select t2.name from friend as t1 join highschooler as t2 on t1.student_id = t2.id;
Question: What are the names and grades for each high schooler?
select name, grade from highschooler;
Question: How many high schoolers are there?
select
\end{lstlisting}
\end{tcolorbox}

\newpage

Below contains an example of a 4-shot cross-domain prompt, which contains 2 demonstration databases, each with 2 demonstration examples ahead of the test database and question. 

\begin{tcolorbox}[title={4-shot cross-domain prompt},breakable, toprule at break=0pt, bottomrule at break=0pt,left=0mm,right=-1mm,top=-1mm, bottom=-1mm,colback=white]
\begin{lstlisting}[style=sql]
create table publication (
publication_id int,
book_id int,
publisher text,
publication_date text,
price real,
primary key (publication_id),
foreign key (book_id) references book(book_id)
);
/*
3 example rows:
select * from publication limit 3;
publication_id    book_id    publisher    publication_date    price
1    1    Pearson    August 2008    15000000.0
2    3    Thomson Reuters    March 2008    6000000.0
3    4    Wiley    June 2006    4100000.0
*/

create table book (
book_id int,
title text,
issues real,
writer text,
primary key (book_id)
);
/*
3 example rows:
select * from book limit 3;
book_id    title    issues    writer
1    The Black Lamb    6.0    Timothy Truman
2    Bloody Mary    4.0    Garth Ennis
3    Bloody Mary : Lady Liberty    4.0    Garth Ennis
*/

-- Using valid SQLite, answer the following questions for the tables provided above.

Question: List the writers of the books in ascending alphabetical order.
select writer from book order by writer asc;
Question: How many books are there?
select count(*) from book;
\end{lstlisting}
\begin{lstlisting}[style=sql]
create table race (
race_id int,
name text,
class text,
date text,
track_id text,
primary key (race_id),
foreign key (track_id) references track(track_id)
);
/*
3 example rows:
select * from race limit 3;
race_id    name    class    date    track_id
1    Rolex 24 At Daytona    DP/GT    January 26 January 27    1
2    Gainsco Grand Prix of Miami    DP/GT    March 29    2
3    Mexico City 250    DP/GT    April 19    2
*/

create table track (
track_id int,
name text,
location text,
seating real,
year_opened real,
primary key (track_id)
);
/*
3 example rows:
select * from track limit 3;
track_id    name    location    seating    year_opened
1    Auto Club Speedway    Fontana, CA    92000.0    1997.0
2    Chicagoland Speedway    Joliet, IL    75000.0    2001.0
3    Darlington Raceway    Darlington, SC    63000.0    1950.0
*/

-- Using valid SQLite, answer the following questions for the tables provided above.
Question: Show the name and location for all tracks.
select name, location from the track;
Question: Show the name of track and the number of races in each track.
select t2.name, count(*) from race as t1 join track as t2 on t1.track_id = t2.track_id group by t1.track_id;

\end{lstlisting}
\begin{lstlisting}[style=sql]

create table highschooler (
id int primary key,
name text,
grade int
);
/*
3 example rows:
select * from highschooler limit 3;
id    name    grade
1510    Jordan    9
1689    Gabriel    9
1381    Tiffany    9
*/

create table friend (
student_id int,
friend_id int,
primary key (student_id,friend_id),
foreign key(student_id) references highschooler(id),
foreign key (friend_id) references highschooler(id)
);
/*
3 example rows:
select * from friend limit 3;
student_id    friend_id
1510    1381
1510    1689
1689    1709
*/

create table likes (
student_id int,
liked_id int,
primary key (student_id, liked_id),
foreign key (liked_id) references highschooler(id),
foreign key (student_id) references highschooler(id)
);
/*
3 example rows:
select * from likes limit 3;
student_id    liked_id
1689    1709
1709    1689
1782    1709
*/

-- Using valid SQLite, answer the following questions for the tables provided above.

Question: How many high schoolers are there?
select
\end{lstlisting}
\end{tcolorbox}

\subsection{Tests of Significance \label{sec:significance}}
Table \ref{tab:schema_prompt_results} contains the performance of Codex and ChatGPT using different database prompt constructions in the zero-shot setting. We observe that the normalization results in slightly improved performance for all database prompt constructions with Codex and 6 out of 7 database prompt constructions with ChatGPT. It is important to note, however, that when comparing normalized and unnormalized database prompt constructions using the same method, the results did not demonstrate statistical significance in McNemar's test, with p-values greater than 0.05. Nevertheless, the primary advantage of normalization lies in its ability to reduce variations among different databases and minimize the overall prompt length.

When evaluating various prompt constructions, we note the advantages gained from incorporating both table relationships (\texttt{Columns=[]+ForeignKey} vs \texttt{Columns=[]}) and table content (\texttt{CreateTable+SelectCol 3} vs \texttt{CreateTable}) are mostly statistically significant in McNemar's test, with p-values smaller than 0.05. Table \ref{tab:schema_prompt_results_significant} displays the results of the significant tests. The performance of \texttt{Columns=[]+ForeignKey} compared to \texttt{Columns=[]} is statistically significant in all cases, except for codex with normalized prompts. Likewise, the performance of \texttt{CreateTable+SelectCol 3} is statistically significant for both Codex and ChatGPT, with both normalized and unnormalized prompts, when compared to \texttt{CreateTable}. These significant findings highlight the effectiveness of incorporating table relationships and database content.

\begin{table*}[!htb] 
    \centering
    \small
    \begin{tabular}{llccc}
    \toprule
    \bf Prompt 1 & \bf Prompt 2 & \bf LLM & \bf Normalization & \bf Significant Test \\
    \midrule
    \multirow{4}{*}{\texttt{Columns=[]}}  & \multirow{4}{*}{\texttt{Columns=[]+ForeignKey}} & Codex & U & \ding{51} \\
    \cmidrule{3-5}
    &  & Codex & N & \ding{55} \\
    \cmidrule{3-5}
    &  & ChatGPT & U & \ding{51} \\
    \cmidrule{3-5}
    &  & ChatGPT & N & \ding{51} \\
    \midrule
    
    \multirow{4}{*}{\texttt{CreateTable}}  & \multirow{4}{*}{\texttt{CreateTable+SelectCol 3}} & Codex & U & \ding{51} \\
    \cmidrule{3-5}
    &  & Codex & N & \ding{51} \\
    \cmidrule{3-5}
    &  & ChatGPT & U & \ding{51} \\
    \cmidrule{3-5}
    &  & ChatGPT & N & \ding{51} \\
    \bottomrule
    \end{tabular}
    \caption{Tests of Statistical Significance for comparing different prompt constructions. Prompt 1 and Prompt 2 were used to represent two distinct methods of constructing prompts in McNemar's test. The prompts were categorized as U and N, representing unnormalized and normalized database prompts, respectively. The \ding{51} symbol indicates that the p-value is smaller than 0.05, indicating statistical significance, while the \ding{55} symbol indicates p-values greater than 0.05, indicating a lack of statistical significance.
    }
    \vspace{-1em}
    \label{tab:schema_prompt_results_significant}
\end{table*}

\subsection{Detailed Single-domain Results}
Tables \ref{tab:single_domain_codex_results} and \ref{tab:single_domain_chatgpt_results} provide detailed results of Codex and ChatGPT in the single-domain setting, respectively. The performance of both models is also illustrated in Figure \ref{fig:single_domain_results}.

\begin{table*}[!htb]
    \centering
    \small
    \begin{tabular}{llccccc}
    \toprule
     \multicolumn{2}{c}{\textbf{Database Prompt Construction}}  & \bf 0-shot & \bf 1-shot & \bf 4-shot & \bf 8-shot & \bf 16-shot \\
    \midrule
    \multirow{2}{*}{\texttt{Table Schema}} & \texttt{Table(Columns)} & 71.9 & 70.7 & 74.9 & 78.0 & 81.6  \\
    \cmidrule{2-7}
    & \texttt{Columns=[]} & 71.8 & 72.1 & 75.5 & 78.0 & 81.6 \\
    \midrule
    \midrule
    \multirow{2}{*}{\texttt{+Relationship}} & \texttt{Columns=[]+ForeignKey} & 73.1&  73.2 & 76.1 & 78.5 & 81.6 \\
    \cmidrule{2-7}
    & \texttt{CreateTable}  & 73.1 & 72.4 & 76.0 & 78.7 & 81.3\\
    \midrule
    \midrule
    \multirow{2}{*}{\texttt{+Relationship+Content}} & \texttt{CreateTable+InsertRow 3} & 71.9 & 72.9 & \underline{77.6} & \underline{80.5} & 82.5\\
    \cmidrule{2-7}
    & \texttt{CreateTable+SelectRow 3}  & \underline{74.1} & \underline{73.4} & 77.3 & \underline{80.5} & \textbf{82.9} \\
    \cmidrule{2-7}
    & \texttt{CreateTable+SelectCol 3}  & \textbf{75.7} & \textbf{74.1} & \textbf{77.9} & \textbf{80.7} & \underline{82.5}\\
    \bottomrule
    \end{tabular}
    \caption{Single-domain results of Codex using different prompt constructions for database schema and content. The best and second-best results for each shot are highlighted in bold and underlined.
    }
    \label{tab:single_domain_codex_results}
\end{table*}

\begin{table*}[!htb]
    \centering
    \small
    \begin{tabular}{llccccc}
    \toprule
     \multicolumn{2}{c}{\textbf{Database Prompt Construction}}  & \bf 0-shot & \bf 1-shot & \bf 4-shot & \bf 8-shot & \bf 16-shot \\
    \midrule
    \multirow{2}{*}{\texttt{Table Schema}} & \texttt{Table(Columns)}  & 70.5 & 71.6 & 74.3 & 77.4 & 79.4  \\
    \cmidrule{2-7}
    & \texttt{Columns=[]}  & 69.1 & 70.7 & 74.4 & 77.8 & 79.5\\
    \midrule
    \midrule
    \multirow{2}{*}{\texttt{+Relationship}} & \texttt{Columns=[]+ForeignKey} & 71.2 & \underline{73.4} & 75.4 & 78.4 & 80.0 \\
    \cmidrule{2-7}
    & \texttt{CreateTable}  & 71.7 & 73.1 & 75.8 & 78.0 & 79.5 \\
    \midrule
    \midrule
    \multirow{2}{*}{\texttt{+Relationship+Content}} & \texttt{CreateTable+InsertRow 3} & 71.8 & 72.8 & \underline{76.6} & \underline{79.1} & \textbf{81.6}\\
    \cmidrule{2-7}
    & \texttt{CreateTable+SelectRow 3} & \underline{72.1} & 73.3 & 76.4 & 78.9 & 81.3 \\
    \cmidrule{2-7}
    & \texttt{CreateTable+SelectCol 3} & \textbf{73.6} & \textbf{73.8} & \textbf{76.8} & \textbf{79.8} & \underline{81.4}\\
    \bottomrule
    \end{tabular}
    \caption{Single-domain results of ChatGPT using different prompt constructions for database schema and content. The best and second-best results for each shot are highlighted in bold and underlined.
    }
    \label{tab:single_domain_chatgpt_results}
\end{table*}

\subsection{Impact of Demonstration Prompt for ChatGPT-16K} \label{sec:appendix:chatgpt-16k}

\begin{figure}[!tb]
  \centering
  \includegraphics[width=.95\linewidth]{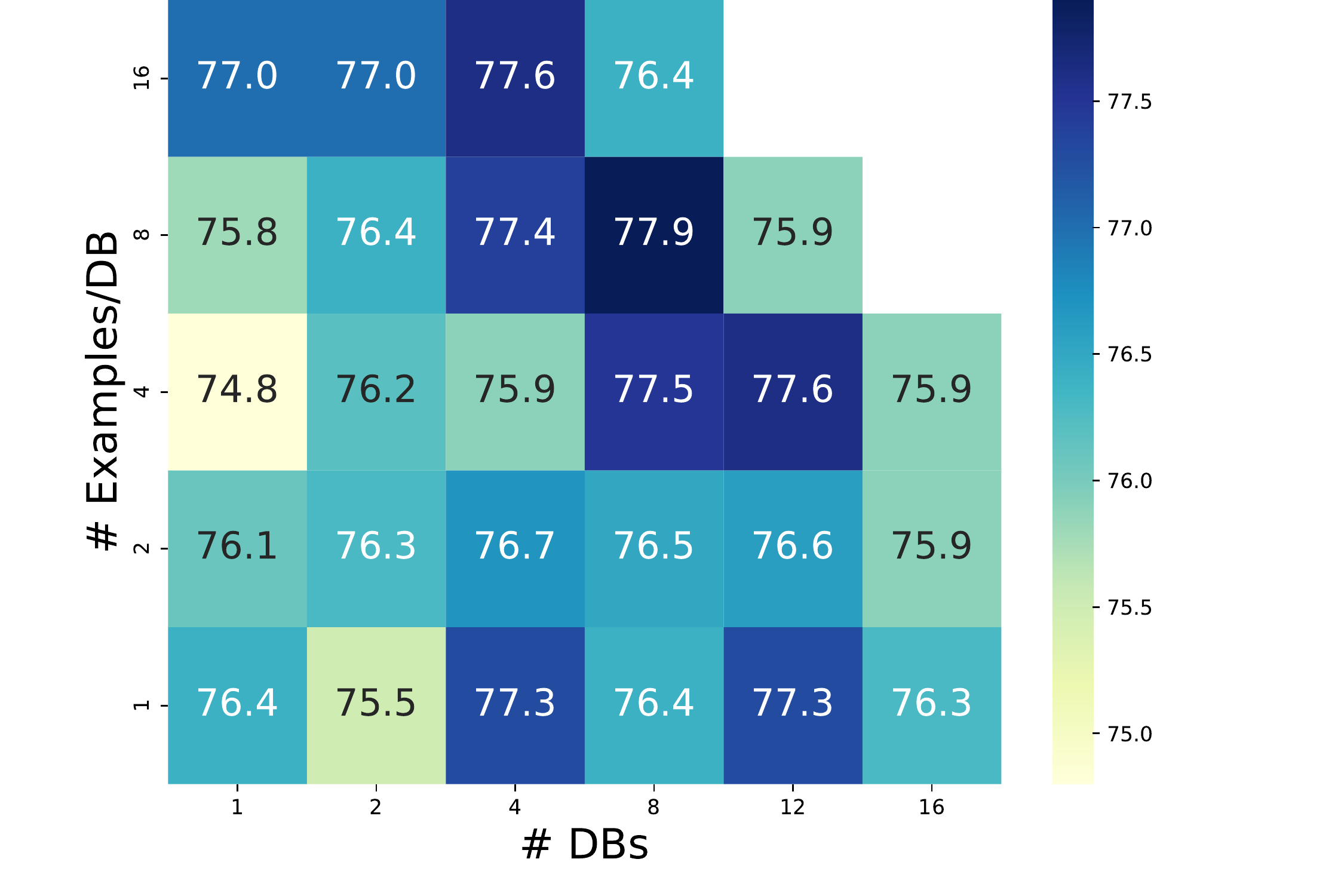}
  \caption{A heat map of ChatGPT-16K's execution accuracy using \texttt{CreateTable+SelectRow 3} for different numbers of databases and examples per database in the demonstration. Darker color indicates higher accuracy.}
  \label{fig:cross_domain_results_chatgpt16k}
  \vspace{-0.5em}
\end{figure}
\begin{figure}[!tb]
  \centering
  \includegraphics[width=.95\linewidth]{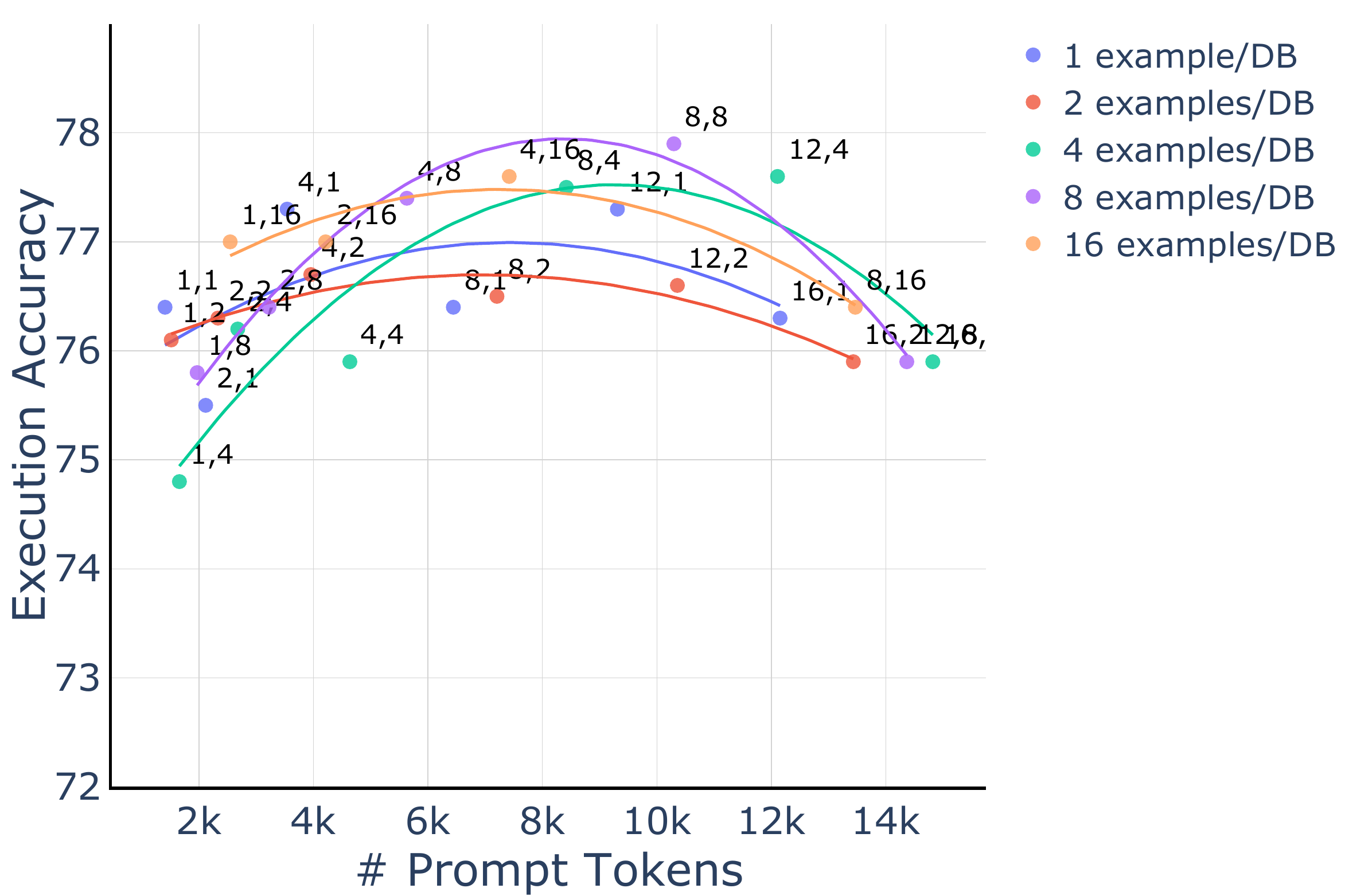}
  \caption{Execution accuracy of ChatGPT-16K in relation to the length of prompts. Each dot represents a demonstration construction, with the $m, k$ denoting the number of databases and examples per database. The lines represent second-degree polynomial trendlines fitted to the results.}
  \label{fig:cross_domain_results_prompt_len_chatgpt16k}
\vspace{-0.5em}
\end{figure}

Figure \ref{fig:cross_domain_results_chatgpt16k} presents the accuracy of ChatGPT-16K corresponding to different combinations of the number of databases and the number of examples per database used as demonstrations.
Similar to our findings with Codex, presenting more databases does not always lead to improved performance for ChatGPT-16K. For a fixed number of examples per database, we observe an initial increase in its performance as the number of databases increases, however, this improvement is followed by a decrease once the number of databases reaches a certain threshold.  
To understand this phenomenon, we analyze the results in relation to the prompt length.

Figure \ref{fig:cross_domain_results_prompt_len_chatgpt16k} shows the relationship between the accuracy of different demonstration prompts and their prompt lengths. Similar to Codex, the performance of ChatGPT-16K also exhibits an inverted-U shape as the prompt length increases for each number of examples per database. Additionally, we observe the performance starts to decrease once the prompt text length exceeds approximately 11K tokens. 

While Codex supports 8K tokens and ChatGPT-16K supports 16K tokens, we notice that their performance tends to decline when dealing with demonstrations that exceed approximately 70\% of the maximum prompt length.

\end{document}